\documentclass[runningheads]{llncs}
\usepackage{booktabs}
\usepackage{float}
\usepackage{pifont}
\usepackage{eccvabbrv}

\usepackage{graphicx}
\usepackage{booktabs}
\usepackage{multirow}
\usepackage{float}

\usepackage[accsupp]{axessibility}  % Improves PDF readability for those with disabilities.

\usepackage{hyperref}

\usepackage{orcidlink}

\begin{document}

\title{TruEye: Fine-Grained Detection of AI-Generated Human Subjects in Images} 

\titlerunning{TruEye: Detecting AI-Generated Human Subjects}


\author{
Jay Barot\inst{1}\orcidlink{0009-0001-5488-6746}
\and
Dan Lin\inst{1}\orcidlink{0000-0003-1521-1452}
}

\authorrunning{J. Barot and D. Lin}

\institute{
Vanderbilt University, Nashville TN, USA\\
\email{jay.d.barot@vanderbilt.edu, dan.lin@vanderbilt.edu}
}

\maketitle

\begin{abstract}
AI generated images are proliferating across the Internet. While some are used for entertainment, others are weaponized for fraud and social engineering attacks on social media users. Existing detectors overfit to generators seen during training, treat detection as opaque binary classification, or rely on costly Large Language Models (LLMs) to explain their outputs. In this paper, we present TruEye, a novel model for fine grained detection and localization of AI manipulated or AI generated humans and scenes. Unlike conventional detectors that assign a single authenticity label, TruEye is the first to distinguish among five compositional categories of synthetic content, including the most challenging case in which a real human is composited into a real scene where they were never physically present. At its core is a mask conditioned dual stream transformer that separates human and scene tokens while preserving patch level spatial correspondence. Specialized reasoning within each stream and region gated cross attention enforce semantic coherence between subject and background, while token level supervision and global compositional classification yield robust, interpretable predictions without invoking an LLM. By restricting intra stream attention to semantically coherent tokens, TruEye also runs over $100\times$ faster than LLM based competitors. Experiments on 6 datasets and our newly curated FineSyn dataset, show that TruEye surpasses state of the art detectors with higher accuracy, faster inference, and stronger generalization to unseen AI generated or manipulated images.
\end{abstract}

\section{Introduction}
\label{sec:intro}

AI-generated images, widely known as deepfakes, are flooding the Internet and have become nearly indistinguishable from authentic photos to the human eye~\cite{microsoft2025realornot,cacm2025coinstoss}. While some are produced for entertainment, others are weaponized for fraud and social engineering at unprecedented scale. ``Pig-butchering'' scams~\cite{abc2024,wired2024}, for instance, now leverage generative AI to fabricate convincing humans and scenes that deceive social media users into transferring funds to attackers. Defending platforms and end users against such attacks requires detectors that satisfy three simultaneous requirements: (i) \emph{generalization} to images produced by generative models unseen during training, (ii) \emph{inference efficiency} sufficient for deployment at platform scale, and (iii) \emph{interpretability}, so that moderators and victims can understand and act on a verdict rather than blindly trust a binary label.
 
Existing detectors fall short on all three fronts. Studies have shown that the accuracy of state-of-the-art detectors degrades sharply when evaluated on images from generative models unseen during training~\cite{abdullah2024analysis}. Most earlier detectors also treat the task as binary classification, i.e., ``real'' versus ``synthetic'', which offers no insight into \emph{what} was manipulated or \emph{where}. Recent detectors, including APSC-net~\cite{miml_apscnet}, SIDA~\cite{huang2025sida}, and LEGION~\cite{kang2025legionlearninggroundexplain}, have begun to provide localization of manipulated regions: some emit pixel-level masks, while others delegate explanation to a Large Language Model (LLM). However, these methods still fail to generalize to unseen or perturbed images, and LLM-driven explanations incur substantial inference latency and compute cost that are prohibitive for content-moderation pipelines processing millions of uploads per day.
 
To address these limitations, we introduce \textbf{TruEye}, a novel detector for fine-grained detection and localization of AI-generated or AI-manipulated humans and scenes. Unlike prior work, TruEye is designed from the ground up around three principles that map directly onto the requirements above.
 
\begin{figure}[!t]
    \centering
    \includegraphics[width=\linewidth]{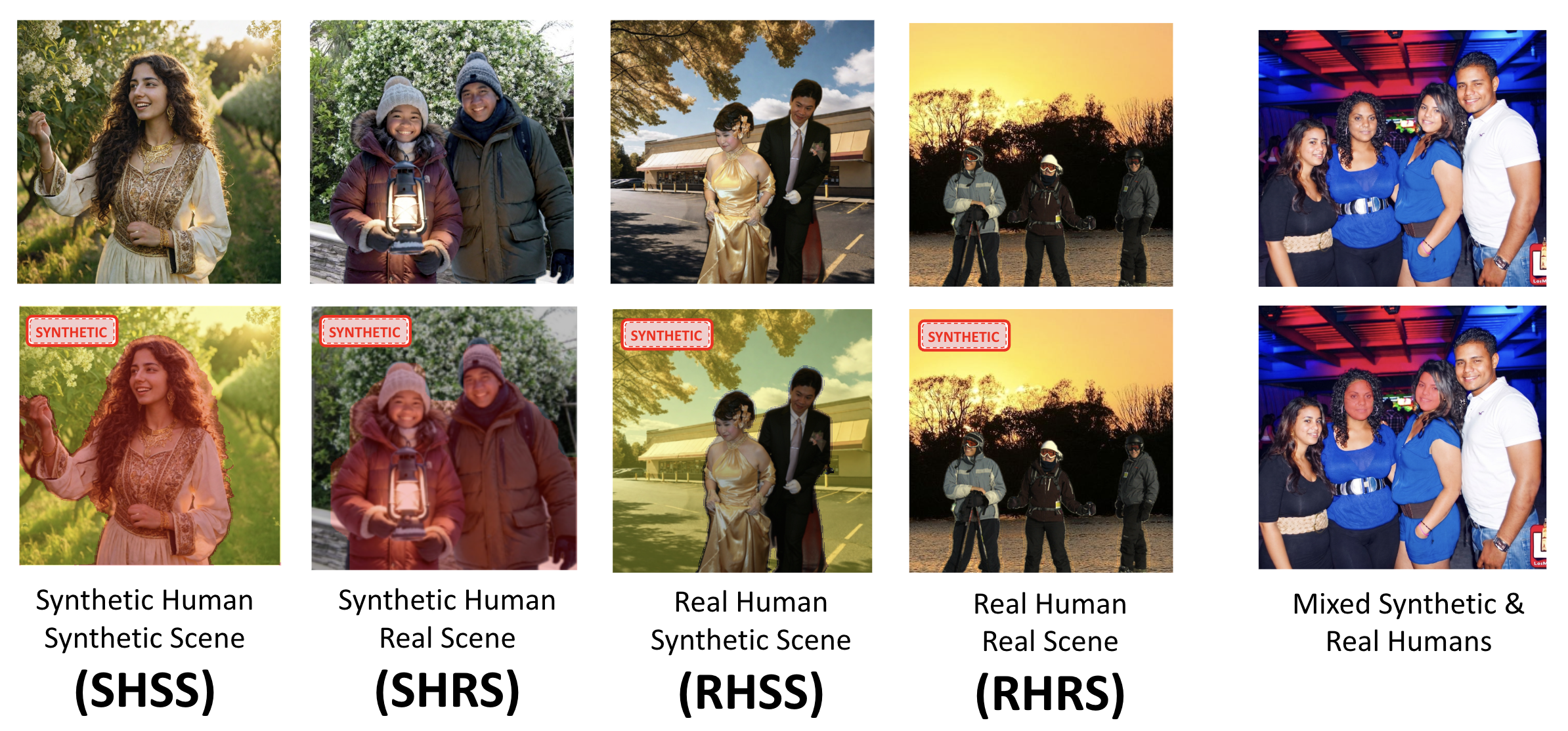}
    \caption{Samples from the FineSyn Dataset and TruEye detection results (labeled).}
    \label{fig:truEye_dataset}
\end{figure}

To the best of our knowledge, TruEye is the first detector that distinguishes among five compositional categories of synthetic content: (a) synthetic human on synthetic scene (SHSS); (b) synthetic human on real scene (SHRS); (c) real human on synthetic scene (RHSS); (d) real human integrated into real scene (RHRS); and (e) synthetic scene (SS). Figure~\ref{fig:truEye_dataset} illustrates how TruEye identifies and classifies manipulated regions. The RHRS category is particularly novel and security-relevant: an authentic human is composited into an authentic scene where they were never physically present. For example, a politician inserted into a protest. Although both the subject and the background are individually real, their co-occurrence is fabricated, making the image manipulative even though no pixel is itself ``synthetic.'' Rather than offloading explanation to an LLM, TruEye produces directly interpretable output: manipulated regions are highlighted through patch-level predictions, and each detection is annotated with a fine-grained category label (e.g., SHRS, RHSS) that conveys \emph{the nature} of the manipulation. This matches the interpretability goals of LLM-based approaches without paying their inference cost.
 
 Naively retrofitting existing detectors with a multi-head classifier does not yield satisfactory fine-grained accuracy, for two reasons. First, most state-of-the-art detectors are architected for binary detection; their feature extractors encode \emph{global} statistical artifacts rather than the \emph{localized} semantic inconsistencies needed to discriminate among categories~\cite{wang2020cnndetection,ojha2023universal}. Second, many existing models implicitly assume that forged regions occupy a substantial fraction of the input image, and consequently perform poorly on RHRS-style manipulations, where the inconsistency is confined to a thin boundary between human and scene~\cite{huh2018selfconsistency,zhang2024cimd}. Even recent localization-aware detectors such as SIDA~\cite{huang2025sida} struggle in this regime.
 
TruEye's key innovation is a \emph{mask-conditioned dual-stream transformer} that decomposes each image into human tokens and scene tokens while preserving patch-level spatial correspondence. Each stream performs domain-specific reasoning, fortified by a magnification module that amplifies subtle artifacts, while a region-gated cross-attention mechanism regulates inter-stream interactions to enforce semantic coherence between subject and background. Token-level supervision aligns every patch's authenticity prediction with its semantic ownership, and a global classification objective enforces compositional consistency across the image. The underlying design philosophy is \emph{divide and conquer}: by decomposing the detection task into specialized subtasks (human--scene separation, human structural analysis, and scene structural analysis), each module concentrates on one objective and specializes in it. This modular design improves learning efficiency, generalizes robustly to unseen generators, and supports localization natively as the final prediction is obtained simply by aggregating patch-level outputs.
 
Beyond accuracy, the dual-stream design substantially reduces inference cost. Rather than allowing every patch to attend to every other patch, human tokens attend only to human tokens and scene tokens attend only to scene tokens, with a single lightweight cross-attention step handling global aggregation. The result is a detector that runs over $100\times$ faster than LLM-based competitors while delivering comparable or richer interpretability. This makes TruEye practical for real-world deployment in content-moderation pipelines, where per-image latency and compute budget are hard constraints.
 
In summary, this paper makes the following contributions:
\begin{itemize}
    \item \textbf{A fine-grained, interpretable detector without LLMs.} We introduce TruEye, the first detector that distinguishes among five compositional categories of synthetic content (SHSS, SHRS, RHSS, RHRS, SS) and produces both localization maps and human-readable category labels \emph{without} invoking a Large Language Model. This delivers the interpretability of LLM-based pipelines at a fraction of the cost.
 
    \item \textbf{A novel mask-conditioned dual-stream architecture.} TruEye's dual-stream transformer, equipped with feature magnification modules and region-gated bi-directional cross-attention, embodies a divide-and-conquer design that specializes one stream for humans and one for scenes. This architecture not only achieves strong fine-grained accuracy---including on the challenging RHRS category---but also restricts intra-stream attention to semantically coherent tokens, yielding over $100\times$ faster inference than LLM-based detectors.
 
    \item \textbf{Extensive evaluation demonstrating generalization and speed.} We compare TruEye against six representative detectors spanning diverse architectural designs, including the 2025 state-of-the-art systems SIDA~\cite{huang2025sida} and LEGION~\cite{kang2025legionlearninggroundexplain}. Evaluations span eight datasets, including OpenForensics~\cite{le2021openforensics}, which was not used to train any of the evaluated models, providing a stringent test of cross-domain generalization. TruEye exhibits substantially stronger generalization while running significantly faster than all baselines.
 
    \item \textbf{The FineSyn dataset.} We curate and release FineSyn, a large labeled dataset of AI-generated images spanning the five fine-grained categories defined above. FineSyn enables the training of TruEye and is intended to support future research on fine-grained synthetic-image analysis and the interpretability of detection results.
\end{itemize}

The remainder of the paper is organized as follows. Section \ref{sec:rel_work} reviews deepfake detection techniques. Section \ref{sec:method} presents our proposed TruEye detector. Section \ref{sec:experiments} reports experimental results. Finally, Section \ref{sec:conclusion} concludes the paper.

\section{Related Work}
\label{sec:rel_work}
As aforementioned, most existing deepfake detectors perform binary classification, labeling an image as either real or fake. A variety of backbone architectures have been explored, including Inception-style networks, Convolutional Neural Networks (CNNs), ResNets, Vision Transformers ViTs and Large Language Models (LLMs) \cite{kong2023vitForgery,dagar2024texvit,huang2025sida,kang2025legionlearninggroundexplain}. In the following, we review these methods in chronological order, along with their underlying architectures.

One of the earliest detectors is XceptionNet~\cite{rossler2019ffpp} which employs an Inception-based architecture with depthwise separable convolutions to extract hierarchical texture cues and leverages ImageNet pre-training for strong cross-dataset generalization. However, its reliance on global texture patterns makes it sensitive to compression and post-processing noise. The Adobe detector~\cite{wang2020cnndetection}, built on a ResNet50 backbone, identifies convolutional artifacts left by early Generative Adversarial Networks (GANs) through residual feature maps. Although it can detect most synthetic images generated by CNN-based generators, it transfers poorly to images created by diffusion  and transformer models.

To enhance generalization across unseen generators, recent approaches incorporate transformer-based attention to capture long-range dependencies \cite{kong2023vitForgery,dagar2024texvit}. For instance, FatFormer~\cite{liu2024fatformer} introduces adaptive adapters that modulate attention heads using forgery priors to enable the model to adjust its receptive field to diverse artifacts.  FakeFormer~\cite{nguyen2024fakeformer} uses vulnerability-guided attention maps that focus on regions prone to manipulation, which enhances robustness under domain shifts. Frequency-based detectors further expose imperceptible inconsistencies. FreqFaceNet~\cite{gupta2025freqfacenet} combines spatial convolution and discrete cosine transform encodings to emphasize high-frequency discrepancies, while frequency-enhanced transformers \cite{tan2024freqaware,wang2023freqresidual,liu2023freqaware} integrate spectral residual cues or frequency-guided positional embeddings to capture subtle generator-specific noise. Hybrid architectures like HTN~\cite{khan2022hybrid} merge convolutional backbones with transformer modules to leverage both local structures and global semantics, and multi-attention models such as MADF~\cite{zhao2021madf} deploy region-specific attention heads to highlight localized inconsistencies in facial areas such as eyes, mouth, and hair. Despite these advances, extending binary detectors to fine-grained multi-class classification remains non-trivial as discussed in the introduction.  This is because when  models are optimized for binary separability in feature space, learned embeddings collapse multiple manipulation modes into a single ``synthetic” representation, which results in poor inter-class discrimination when applied to more nuanced categories as defined in this work.

Recent works have also explored richer manipulation understanding beyond binary classification. MIML/APSC-Net~\cite{miml_apscnet} adopts a multi-instance multi-label framework to jointly predict multiple manipulation attributes, enabling more detailed semantic reasoning. GIMFormer~\cite{gimformer} employs transformer-based representations to model manipulation presence, category, and affected regions in a unified framework. Most recently, OpenSDI/MaskCLIP~\cite{wang2025opensdid} leverages CLIP-based vision--language representations with mask-guided supervision to improve both synthetic image detection and artifact localization. While these methods move beyond conventional binary detection, they do not explicitly address the fine-grained semantic categories considered in our work.

Closely related to our proposed fine-grained detection, several localization-oriented forensic models attempt to identify manipulated regions at the pixel or patch level. ManTra-Net~\cite{wu2019mantra} formulates detection as dense anomaly estimation by employing encoder–decoder architectures to highlight local inconsistencies between predicted and reconstructed patches.  Face X-ray~\cite{li2020facexray} focuses on facial composites by detecting blending boundaries via gradient-domain analysis, effectively capturing transitions between real and synthetic facial regions. MVSS-Net~\cite{chen2021mvss} introduces multi-scale feature extraction and boundary supervision to better detect splicing, copy-move, and removal artifacts by enforcing edge-level consistency constraints. TruFor~\cite{guillaro2023trufor} fuses RGB and noise fingerprints using a transformer-based multimodal fusion module. The most recent LLM-based detector, SIDA ~\cite{huang2025sida}, introduces  a multimodal framework that jointly leverages visual and textual representations for spatially grounded forgery detection. It employs cross-modal attention to correlate localized regions with language-based cues and integrates pixel-level segmentation heads for artifact localization \cite{zhou2024sig}. Similarly, LEGION~\cite{kang2025legionlearninggroundexplain} proposes a multimodal large language model based framework that performs forgery detection, artifact localization, and textual explanation. It uses both visual and textual information to produce segmentation masks together with artifact descriptions. It explores using artifact explanations as feedback to guide synthetic image refinement and regeneration. However, none of these localization methods provide fine-grained semantic reasoning like our proposed TruEye, which explicitly distinguishes whether the synthetic component involves the human, the scene, or both. In our experiments, we systematically compare our approach with these methods and demonstrate that our model achieves superior detection accuracy and stronger generalizability.

\section{The Proposed TruEye Detector}
\label{sec:method}

In this section, we present \textbf{TruEye}, a novel fine-grained detector for identifying AI-generated human subjects and scenes in images.

\subsection{Architecture Overview}
\begin{figure}[!t]
    \centering
    \includegraphics[width=\linewidth]{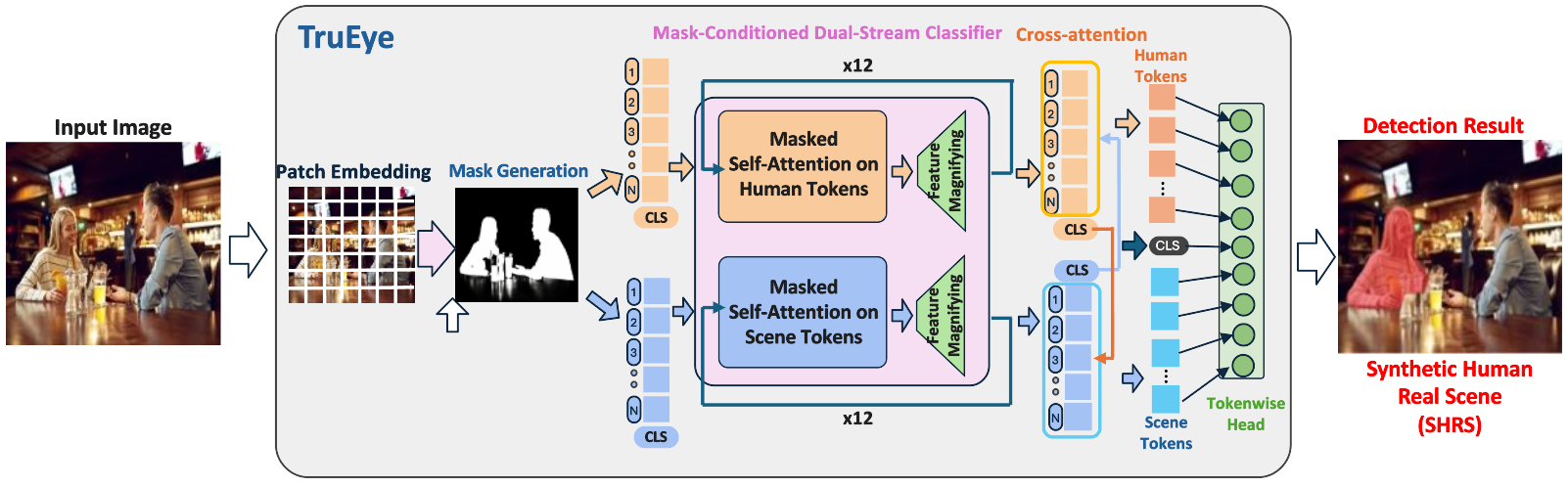}
    \caption{
    An Overview of TruEye Architecture. (The depicted input image, predicted authenticity map, and patch numbering are included solely for demonstration purposes.)
    }

    \label{fig:trueye_architecture}
\end{figure}

TruEye has a unique \emph{mask-conditioned dual-stream architecture} built upon transformer blocks.  Figure \ref{fig:trueye_architecture} illustrates the overall architecture of TruEye which consists of four main components: (i) mask generator; (ii) dual-stream classifiers;  (iii) feature magnifier; and (iv) global cross attention layer.

Given an input image \(x \in \mathbb{R}^{3 \times H \times W}\), TruEye will output a prediction label in one of the following six categories: (a) synthetic human on synthetic scene (SHSS); (b) synthetic human on real scene (SHRS); (c) real human on synthetic scene (RHSS); (d) real human integrated into real scene (RHRS); (e) synthetic scene (SS); and (f) fully real image (R). Specifically, an image will be processed as follows. 

% \begin{equation}
%   E = \mathrm{Conv}_{P}(x)
%     \in \mathbb{R}^{B \times D \times \frac{H}{P} \times \frac{W}{P}}
%   \label{eq:patch_embedding}
% \end{equation}
%where $B$ denotes the batch size. After flattening and reshaping, we obtain a token sequence \(t \in \mathbb{R}^{B \times N \times D}\), where \(t\) is obtained by flattening the spatial dimensions of \(E\). This sequence is then augmented with a learnable global classification token [CLS] 

\paragraph{\bf Input Embedding.} We first divide the input image $x$ into non-overlapping $P \times P$ patches. Each patch is linearly projected into a $D$-dimensional embedding space using a convolutional layer with kernel size $P$ and stride $P$, which serves as the patch embedding layer. In our experiments, images are partitioned into $16 \times 16$ patches, yielding $N=196$ patch tokens. A learnable global classification token $[\text{CLS}]$ is prepended to the sequence of patch embeddings. Positional encodings are added element-wise to token embeddings and have the same dimensionality $D$ as the token representations. The resulting input sequence $x_0$ is defined as:
\begin{equation}
x_0 = [c; t] + \mathbf{p},
\label{eq:token_with_pos}
\end{equation}
where $c \in \mathbb{R}^{D}$ denotes the $[\text{CLS}]$ token, $t \in \mathbb{R}^{N \times D}$ represents the sequence of patch embeddings, and $\mathbf{p} \in \mathbb{R}^{(N+1) \times D}$ is the positional encoding. 

\paragraph{\em\bf Mask Generator.} The TruEye's mask generator predicts a binary mask $m$ for each image token representation (excluding the [CLS] token), where $m = 1$ indicates that more than 50\% of the corresponding image patch contains human-related content, and $m = 0$ indicates that the patch is predominantly scene content.  The mask generator is implemented as one transformer layer. The design of this component serves two main purposes. First, it decomposes the artificial artifact detection task into smaller subtasks, enabling the subsequent dual-stream classifier to focus on specific content types, either human or scene. Second, the mask generator allows the model to handle information regarding arbitrary human-to-scene ratios without modifying the overall token structure as described below.

\paragraph{\bf Dual-Stream Classifier.} After mask generation, image tokens are fed into a dual-stream classifier composed of two parallel branches, namely the human stream and the scene stream. Each stream contains 12 stacked layers, where each layer consists of a Transformer block followed by a Feature Magnification Module (FMM). To preserve compatibility with pretrained ViT weights, both streams operate on the full set of image tokens instead of removing tokens associated with the complementary region. Region-specific feature refinement is controlled via masking, such that in the human stream only human-region tokens are active, whereas in the scene stream only scene-region tokens are active. Specifically, to ensure that  active tokens are refined exclusively through intra-region interactions and that inactive tokens neither influence nor are influenced by the refinement process, attention logits are modified prior to softmax by adding a masking matrix $\mathcal{M}$ such that attention is permitted only between active tokens, while all interactions involving inactive tokens are suppressed. To further prevent unintended updates, we apply a region-gated residual mechanism. Specifically, residual updates from both the self-attention and MLP sublayers are applied only to active tokens, while inactive tokens bypass these updates and remain unchanged.  This design maintains architectural compatibility with pretrained ViT blocks, preserves fixed tensor dimensions for efficient batching, and enforces strict region-specific feature learning without cross-region interference.

\paragraph{\bf Feature Magnification Module (FMM).} The Feature Magnification Module  is a two-layer fully connected bottleneck after each transformer block. It first projects the output from the Transformer block from  768 dimensions to 48 dimensions and then expands back to 768, enforcing a compressed intermediate representation that promotes informative feature learning. The resulting representation is converted into channel-wise scaling factors, namely excitation factors, constrained to  $[0.5, 1.5]$. Each channel of the original token feature is then multiplied by its corresponding excitation factor to amplify important channels (scale $>1$) and suppressing less informative ones (scale $<1$). In this way, FMM magnifies discriminative cues within human or scene tokens, which in turn improves sensitivity to localized inconsistencies.  Formally, the output from FMM is represented as follows: 
\begin{equation}
  x \leftarrow \left(0.5 + \sigma(g(x))\right) \odot x,
  \label{eq:feature_magnifying}
\end{equation}
where $g(x)$ is the output from the two-layer bottleneck, $\sigma(\cdot)$ is the sigmoid function and $\odot$ denotes channel-wise multiplication.

% To jointly capture patch-level anomalies and image-wide contextual inconsistencies, TruEye adopts a hierarchical supervision strategy that aligns learning objectives across semantic scales. At the token level, the model learns localized authenticity cues by directly associating each patch representation with region-specific labels derived from the human–scene mask composition. At the global level, the [CLS] representation aggregates contextual information across all regions, enabling the network to assess overall image consistency and distinguish composites from fully authentic images.

\paragraph{\bf Bidirectional Cross-Attention on [CLS] Tokens.} After the 12 stacked Transformer + FMM blocks, we introduce a cross-attention mechanism between the [CLS] tokens of the two streams. Since each [CLS] token aggregates contextual information over all regions within its respective stream, this cross-attention operation enables the model to assess global image consistency. 
Specifically, the [CLS] token from the human stream attends to all image tokens in the scene stream, while the [CLS] token from the scene stream attends to all image tokens in the human stream, as formulated below:
\begin{equation}
\begin{aligned}
h_0&\leftarrow h_0 + \phi(h_0, s, k^{(s)}), ~~~
s_0&\leftarrow s_0 + \phi(s_0, h, k^{(h)})
\end{aligned}
\label{eq:cross_attention}
\end{equation}
where \(h_0\) and \(s_0\) denote the [CLS] tokens of the human and scene streams, respectively, and \(\phi\) is a masked cross-attention operator.  
This lightweight interaction encourages mutual alignment without incurring full inter-token computation. This design is also critical for distinguishing the most challenging category ``Real Human Real Scene'' from fully real images. Patch-level predictions alone cannot differentiate the two as all patches would be classified as real in both cases.

\paragraph{\bf Final Output.} After cross-attention, the two updated $[\text{CLS}]$ tokens are fused into a single global token. We retain the human tokens from the human stream and the scene tokens from the scene stream, yielding 197 tokens in total (one global token and 196 patch tokens). A shared linear layer is applied to each token to produce an authenticity logit, which is converted to a probability via a sigmoid function. Probabilities closer to 1 indicate the ``synthetic'' class, while those closer to 0 indicate ``real''.  

\paragraph{\bf Inference.} During inference, a binary decision is made by thresholding the output probability at 0.5. If at least one human token is classified as ``synthetic'',  the image is labeled as Synthetic Human (SH). Similarly, if at least one scene token is classified as ``synthetic'', the image is labeled as Synthetic Scene (SS). The $[\text{CLS}]$ token is used to distinguish ``Real Human Real Scene" category from fully real images. An image is predicted as ``real'' only when the $[\text{CLS}]$ token is classified as real and no human or scene token is predicted as synthetic. Finally, localization is naturally derived from the patch-level prediction results, which indicate which regions of the image are synthetic.

% predicts a token-level authenticity map:
% \begin{equation}
%   f_{\theta}(x, m) = \{ z_i \}_{i=0}^{N}, \quad z_i \in \mathbb{R}
%   \label{eq:token_map}
% \end{equation}
% where \(N = (H / P) \cdot (W / P)\) represents the number of non-overlapping \(P \times P\) patches.  
% Each token logit \(z_i\) quantifies the probability that the corresponding patch is AI-generated (\(1\)) or real (\(0\)). 

\subsection{Two-Stage Training}

TruEye is optimized in two stages to stabilize learning and encourage hierarchical supervision. During training, we initialize the model with pretrained ViT weights,  set the initial learning rate to $3 \times 10^{-4}$, and use the AdamW optimizer with a mini-batch size of 32. The model is trained for 138 epochs with early stopping applied to prevent overfitting.

\paragraph{\bf Stage I: Mask Pretraining.}
We first train the mask generator independently using patch-level binary supervision:
\begin{equation}
\mathcal{L}_{\text{msk}} 
= \frac{1}{N} \sum_{i=1}^{N}
\mathrm{BCE}_\epsilon\bigl(\sigma(m_i),\, y^{(\text{msk})}_i \bigr),
\label{eq:mask_loss}
\end{equation}
where $m_i$ denotes the predicted logit for the $i$-th patch, $\sigma(\cdot)$ is the sigmoid function, $y^{(\text{msk})}_i \in \{0,1\}$ is the ground-truth binary mask label, and $N$ is the number of patches. $\mathrm{BCE}_\epsilon$ denotes numerically stable binary cross-entropy.

This stage provides strong localized supervision. Empirically, this initialization improves optimization stability and prevents the global objective from dominating early training.

\paragraph{\bf Stage II: End-to-End Optimization.}
We then fine-tune the entire TruEye framework end-to-end using a multi-level objective:
\begin{equation}
\mathcal{L}
= \mathcal{L}_{\text{msk}} 
+ \mathcal{L}_{\text{tok}}(w_{\text{pos}})
+ \lambda_{\text{cls}} \, \mathcal{L}_{\text{cls}},
\qquad \lambda_{\text{cls}} = 0.6,
\label{eq:total_loss}
\end{equation}

where $\lambda_{\text{cls}}$ balances global and local supervision. The token-level loss is defined as
\begin{equation}
\mathcal{L}_{\text{tok}}
= \frac{1}{N} \sum_{i=1}^{N}
\mathrm{BCE}_\epsilon\bigl(\sigma(z_i),\, y^{(\text{tok})}_i;\, w_{\text{pos}} \bigr),
\label{eq:token_loss}
\end{equation}

where $z_i$ denotes the token-level logit and $\epsilon$ is the label smoothing parameter. We set $\epsilon = 0.02$ to mitigate overconfidence and improve generalization. Also, to address potential class imbalance, we apply a dynamic positive weighting $w_{\text{pos}}$ in the token-level loss $\mathcal{L}_{\text{tok}}$
\begin{equation}
w_{\text{pos}}
= \mathrm{clip}\!\left(
\frac{N_{\text{neg}}}{N_{\text{pos}}},
0.5,\, 2.0
\right),
\label{eq:pos_weight}
\end{equation}
where $N_{\text{pos}}$ and $N_{\text{neg}}$ denote the numbers of synthetic and real tokens in the batch, respectively. This weighting increases the penalty for misclassifying synthetic tokens. When a batch contains few synthetic tokens, their loss is upweighted to prevent the model from minimizing loss by predicting the majority class (real). This keeps gradient contributions balanced across categories. The weight is clipped to ensure stable updates. Finally, the global token loss is
\begin{equation}
\mathcal{L}_{\text{cls}} 
= \mathrm{BCE}\bigl(\sigma(z_0),\, y_{\text{cls}}\bigr),
\qquad y_{\text{cls}} \in \{0,1\},
\label{eq:cls_loss}
\end{equation}
where $z_0$ is the fused $[\text{CLS}]$ logit and $y_{\text{cls}}$ indicates whether the image is synthetic.
The global supervision encourages the $[\text{CLS}]$ embedding to capture high-level semantic consistency and cross-region dependencies that may not be fully represented by patch-level gradients.

% For each training image, two binary indicators \((y_h, y_s) \in \{0,1\}^2\) describe the authenticity of the human and scene components, respectively.  
% The patch-level ownership map \(p\), obtained during region-conditioned factorization, specifies whether each token corresponds to a human or scene region.  
% We construct a token-level label map as:
% \begin{equation}
%   y^{(\text{tok})} = p\,y_h + (1 - p)\,y_s
%   \label{eq:token_labels}
% \end{equation}
%assigning to each patch token the authenticity of its semantic region.  
%Let \(\{z_i\}_{i=1}^{N}\) denote the predicted patch-level logits from the fused token representation.  
%These logits are optimized using a smoothed binary cross-entropy objective:

%This hierarchical formulation establishes bidirectional coupling: token-level losses refine spatial discrimination of subtle forgeries, while global supervision regularizes contextual reasoning.  Together with the region-conditioned dual-stream transformer, this synergy promotes coherent, interpretable authenticity maps that maintain semantic consistency across human–scene boundaries.

\subsection{FineSyn Dataset}
\label{sec:finesyn_dataset}
To enable systematic study of cross-domain authenticity cues between real and synthetic human–scene compositions, we created \textit{FineSyn}, a curated dataset for fine-grained compositional analysis. To the best of our knowledge, \textit{FineSyn} is the first dataset that explicitly covers all combinations of real and synthetic humans and scenes. Each synthetic image was generated under manual supervision to ensure visual and semantic consistency. FineSyn focuses on human-centric imagery and spans diverse subjects, environments, and lighting conditions. We used an 80–20 split for training and evaluation.

We structured FineSyn into six fine-grained categories as described in Section 3.1. We generated 5,000 images for each synthetic composition type and curated 10,000 diverse real human-centric images to strengthen authenticity priors and help the model learn clear distinctions between real and synthetic content. In total, FineSyn contains 35,000 images that comprehensively represent the full spectrum of human–scene compositions. Figure~\ref{fig:truEye_dataset} illustrates representative samples from categories involving human subjects and TruEye’s detection results. Synthetic-human regions are highlighted in red, and synthetic-scene regions are highlighted in yellow. For samples belonging to the real-human–real-scene category, TruEye produces no regional highlights and instead outputs the category label RHRS.

We built the FineSyn dataset using a structured multi-stage pipeline that combines diffusion and GAN-based synthesis~\cite{rombach2022stablediffusion,karras2019stylegan2}, segmentation, manual composition, and harmonization. For diffusion-based generation, we used Stable Diffusion~\cite{rombach2022stablediffusion} with complementary levels of prompt diversity control: a scene-level framework varying location type, lighting, camera viewpoint, and weather, and a human-level framework adjusting age, clothing, pose, ethnicity, and other visual attributes. For GAN-based synthesis, we adopted multiple pretrained models, including GauGAN, StarGAN, CycleGAN, BigGAN, StyleGAN, ProGAN, and StyleGAN2~\cite{park2019gaugan,choi2018stargan,zhu2017cyclegan,brock2018biggan,karras2019stylegan,karras2018progressive,karras2019stylegan2} to ensure visual and stylistic diversity across human and scene distributions. To incorporate real content, we collected humans and backgrounds from COCO, ImageNet, FFHQ, CelebA-HQ, and Human Pose datasets~\cite{lin2014coco,deng2009imagenet,karras2019ffhq,karras2018progressive,andriluka2014pose}. 

We segmented both real and synthetic humans and composed them into real or synthetic scenes using a custom web application that enabled precise spatial control. For categories such as SHSS, we utilized generative models capable of producing coherent human–scene composites directly, minimizing manual intervention. After composition, we refined each image using a harmonizer model~\cite{ke2022harmonizer} to standardize illumination, color balance, and texture.

% \subsection{Training Details}
% We train TruEye for 120 epochs using AdamW~\cite{loshchilov2019adamw} (learning rate $3\times10^{-4}$, weight decay $0.05$) with a 15-epoch warmup and cosine annealing~\cite{loshchilov2017sgdr}. Training uses batch size 64, mixed precision, gradient clipping at 1.0, and an exponential moving average of weights~\cite{tarvainen2017meanteacher} (decay $0.999$) for evaluation. The loss combines token-level binary cross-entropy with label smoothing~\cite{szegedy2016rethinking} $0.02$ and adaptive positive weighting, plus a CLS term weighted by $\lambda_{\text{cls}}=0.6$. Each patch token inherits its label from the corresponding human or scene region. We apply random flips and color jitter for augmentation. Early stopping triggers when validation loss fails to improve for 20 epochs within a threshold of $5\times10^{-4}$. We use 80\% of the FineSyn dataset for training and  the remaining 20\%  for testing. 

\section{Experiments}
\label{sec:experiments}

All experiments were conducted on a workstation equipped with an Intel Core i9-14900KF CPU (32 cores), 62 GB RAM, and a single NVIDIA GeForce RTX 4090 GPU (24 GB).  

For a comprehensive evaluation, we compare TruEye with three representative detectors featuring diverse architectures, including classical convolutional networks, modern transformers, and recent LLM-based models, as discussed in Section~\ref{sec:rel_work}.  Specifically, we compared our model with two 2025 models, SIDA~\cite{huang2025sida} and LEGION~\cite{kang2025legionlearninggroundexplain}.  It is worth noting that both SIDA and LEGION do not explicitly classify manipulated images into the fine-grained categories defined in this paper. SIDA has already been evaluated against many prior detectors, we further selected another representative model TruFor~\cite{guillaro2023trufor} which has not been compared with SIDA yet. Table~\ref{tab:model_size} summarizes each model's publication venue, year, base architecture, number of parameters, and model size.

The datasets used for comparison include most of the training datasets released with the models evaluated in this study. In addition, we include external datasets such as OpenForensics \cite{le2021openforensics}, FaceForensics++ \cite{rossler2019ffpp} and OpenSDID \cite{wang2025opensdid} which is not used for training by any of the evaluated models, including our TruEye, to provide an unbiased test set. Unlike many earlier deepfake datasets that focus on single-face images, OpenForensics includes diverse in-the-wild scenes with multiple faces, making it a more challenging benchmark for evaluating manipulation detection methods. FaceForensics++ is one of the most widely adopted benchmarks in digital media forensics and contains high-qualit manipulations generated using multiple face forgery techniques, making it valuable. OpenSDID provides a large-scale and balanced collection of real and AI-generated images produced by multiple state-of-the-art diffusion models, offering a challenging benchmark for evaluating the generalization capability of synthetic image detection methods across diverse generative sources.

\begin{table}[!t]
\centering
\scriptsize
\caption{Model Structure Type, Size and Capabilities}
\label{tab:model_size}
\setlength{\tabcolsep}{4pt}
\renewcommand{\arraystretch}{0.9}
\begin{tabular}{lccc}
\toprule
Model  & Base Architecture  & \# Params & Size  \\
\midrule
TruFor [CVPR 2023] & MiT-B2 + DnCNN & 68.7M & 255.9 MB  \\
LEGION [ICCV 2025] & LLM + ViT + SAM & 8B & 29.8 GB \\
SIDA [CVPR 2025] & LLM & 13B & 28.8 GB  \\
\textbf{TruEye (ours)} & \textbf{ViT} & {\bf 263M} & \textbf{1 GB} \\
\bottomrule
\end{tabular}
\end{table}

\begin{table}[!t]
\centering
\scriptsize
\caption{Summary of Datasets}
\label{tab:data}
\setlength{\tabcolsep}{4pt}
\renewcommand{\arraystretch}{0.9}
\begin{tabular}{l|c|c|c|c}
\toprule
\textbf{Dataset} & \textbf{Year} & \textbf{\# Synthetic Images} & \textbf{\# Real Images} & \textbf{Model Trained On It} \\
\midrule

FaceForensics++    & 2019 & 10,000 & 1363 & XceptionNet \\

OpenForensics      & 2021 & 95,134 & 95,201 & None \\

SID-set            & 2025 & 200,000 & 100,000 & SIDA \\

OpenSDID            & 2025 & 150,000 & 150,000 & MaskCLIP \\

SynthScars         & 2025 & 12,236 & 0 & LEGION \\

FineSyn (Ours)     & 2026 & 25,000 & 10,000 & TruEye \\
\bottomrule
\end{tabular}
\end{table}

 %comparison and a FaceForensics++ (FF+) \cite{rossler2019ffpp} by XceptionNet, ForenSynth \cite{wang2020cnndetection} by the Adobe Detector, CASIA, FantasticReality \cite{} by TruFor
%SID-Set \cite{huang2025sida} by the SIDA detector,  and our newly curated FineSyn dataset. FaceForensics++ (FF+) contains 1,000 pristine and manipulated videos generated with DeepFakes, Face2Face, FaceSwap, and NeuralTextures, which allows controlled evaluation of human-centric forgeries. The ForenSynth dataset contains over one million GAN-generated images produced by ProGAN, StyleGAN, BigGAN, and CycleGAN. SID-Set includes approximately 300,000 images created by diffusion and GAN-based methods and divides them into real, synthetic, and tampered categories, where the synthetic category refers to fully generated images and the tampered category refers to partially manipulated ones. 

For a fair comparison and rigorous evaluation of generalization capability, all compared detectors are employed with their {\bf originally published parameters}; and our TruEye model is trained exclusively on our newly curated FineSyn dataset which contains no overlap with the data used to train other detectors. 

\subsection{Performance of Fine-grained Analysis}

In the first round of experiments, we evaluate the detection effectiveness of TruEye on the FineSyn dataset. Since TruEye is built upon vision transformers, we also trained a standard vision transformer equipped with a multi-head classifier as a baseline for comparison. As shown in Table \ref{tab:trueye_vit_finesyn}, TruEye achieves an overall accuracy of 95.52\% and F1-score 97.08\%, whereas the pure vision transformer reaches only 60.11\% accuracy and 53.31\% F1-score. Moreover, TruEye performs exceptionally well in the most challenging category (RHRS) where a real human is integrated into a real scene. This result indicates that simply augmenting a vision transformer with a multi-head classifier is insufficient for performing fine-grained analysis tasks. 

\begin{table}[!ht]
\centering
\caption{Performance of TruEye \& Pure ViT on FineSyn dataset}
\small
\resizebox{\textwidth}{!}{
\begin{tabular}{lcccccccccccccc}
\toprule
\multirow{2}{*}{\textbf{Model}} & 
\multicolumn{2}{c}{\textbf{SHSS}} & 
\multicolumn{2}{c}{\textbf{SHRS}} & 
\multicolumn{2}{c}{\textbf{SS}} & 
\multicolumn{2}{c}{\textbf{RHSS}} & 
\multicolumn{2}{c}{\textbf{RHRS}} & 
\multicolumn{2}{c}{\textbf{Real}} & 
\multicolumn{2}{c}{\textbf{Overall}} \\
\cmidrule(lr){2-3} 
\cmidrule(lr){4-5} 
\cmidrule(lr){6-7}
\cmidrule(lr){8-9}
\cmidrule(lr){10-11} 
\cmidrule(lr){12-13} 
\cmidrule(lr){14-15}
 & IoU(\%) & F1(\%)
 & IoU(\%) & F1(\%)
 & IoU(\%) & F1(\%)
 & IoU(\%) & F1(\%)
 & IoU(\%) & F1(\%)
 & IoU(\%) & F1(\%)
 & IoU(\%) & F1(\%)\\
\midrule
\textbf{TruEye}
& \textbf{90.34} & \textbf{94.93}
& \textbf{89.93} & \textbf{94.70}
& \textbf{93.05} & \textbf{96.40}
& \textbf{94.35} & \textbf{97.09}
& \textbf{97.13} & \textbf{98.54}
& \textbf{97.43} & \textbf{98.70}
& \textbf{93.70} & \textbf{96.73} \\

Pure ViT
& 78.69 & 85.22
& 24.14 & 27.96
& 80.26 & 86.19
& 79.14 & 83.17
& 77.59 & 81.50
& 75.89 & 77.60
& 69.29 & 73.61 \\
\bottomrule
\end{tabular}
}
\label{tab:trueye_vit_finesyn}
\end{table}

\subsection{Comparison with State-of-the-art Detectors}

The primary objective of these experiments is to assess the generalization performance of TruEye compared with existing detectors and their abilities to satisfy real-time deployment requirements measured  by per-image inference latency.

As aforementioned, each detector is employed with its originally published parameters to eliminate potential discrepancies arising from retraining. Likewise, our TruEye model is trained solely on the FineSyn dataset. Because existing detectors do not perform fine-grained classification across the five synthetic categories defined in this study, we consolidate these fine-grained categories in our FineSyn dataset under a single ``synthetic'' label when testing other detectors. Similarly, we treat TruEye's detection results of any of the five synthetic categories as ``synthetic'' when comparing to other detectors. 

\renewcommand{\arraystretch}{0.9}
\begin{table*}[!t]
\centering
\caption{Generalization Performance, report AUC and IoU.}
\label{tab:cross_generalization_all}
\resizebox{\textwidth}{!}{%
\begin{tabular}{lcccccc|cccccc}
\toprule
\textbf{Method}
& \multicolumn{2}{c}{\textbf{SID-set}}
& \multicolumn{2}{c}{\textbf{SynthScars}}
& \multicolumn{2}{c}{\textbf{FineSyn}}
& \multicolumn{2}{|c}{\textbf{\textcolor{blue}{FF++}}}
& \multicolumn{2}{c}{\textbf{\textcolor{blue}{OpenForensics}}}
& \multicolumn{2}{c}{\textbf{\textcolor{blue}{OpenSDID}}}
\\

& \multicolumn{2}{c}{\small(2025)}
& \multicolumn{2}{c}{\small(2025)}
& \multicolumn{2}{c}{\small(2026)}
& \multicolumn{2}{|c}{\small(2019)}
& \multicolumn{2}{c}{\small(2021)}
& \multicolumn{2}{c}{\small(2025)}
\\

\cmidrule(lr){2-7}
\cmidrule(lr){8-13}

& AUC(\%) & IoU(\%)
& AUC(\%) & IoU(\%)
& AUC(\%) & IoU(\%)
& AUC(\%) & IoU(\%)
& AUC(\%) & IoU(\%)
& AUC(\%) & IoU(\%) \\
\midrule

\textit{TruFor}
& 74.08 & 35.00
& n.a & 3.65
& 78.78 & 33.60
& 62.93 & 54.79
& 80.49 & 71.39
& 55.08 & 33.91 \\

\textit{SIDA}
& \textbf{87.30} & \textbf{73.90}
& n.a & 60.40
& 54.92 & 48.28
& 72.36 & 48.39
& 58.88 & 60.25
& 58.88 & 8.25 \\

\textit{LEGION}
& 71.74 & 35.36
& n.a & \textbf{55.91}
& 47.43 & 33.81
& 77.88 & 50.30
& 8.32 & 4.26
& 64.57 & 41.17 \\

\textit{\textbf{TruEye}}
& \boxed{96.50} & \boxed{79.82}
& n.a & \boxed{61.58}
& \boxed{\textbf{98.32}} & \boxed{\textbf{93.70}}
& \boxed{89.23} & \boxed{82.96}
& \boxed{93.20} & \boxed{83.46}
& \boxed{91.85} & \boxed{79.82} \\

\bottomrule
\end{tabular}%
}
\end{table*}

\subsubsection{Detection Accuracy}

Table~\ref{tab:cross_generalization_all} presents the cross-dataset evaluation results of all  detectors across the 6 datasets. For each method, we report AUC and F1-score. The numbers highlighted in bold indicate the dataset on which the corresponding model was originally trained, and the boxed numbers indicate the best performer for that dataset.

As shown in the table, most existing methods achieve their best performance on the datasets used for training but experience noticeable performance drops when evaluated on other datasets. This indicates their limited cross-dataset generalization.

In contrast, the proposed TruEye achieves consistently strong performance across all datasets, which demonstrates its superior robustness to dataset shifts. Unlike other models that peak only on their training datasets, TruEye maintains high accuracy and F1 scores across multiple benchmarks, ranking as the top performer on all datasets.

Notably, OpenForensics, FaceForensics++ and OpenSDID are dataset on which none of the models were trained, and hence provides a strict evaluation of generalization ability. On this unseen dataset, TruEye outperforms all competing methods. TruEye’s strong generalization ability likely stems from its divide-and-conquer design philosophy, where each module specializes in a smaller subtask, combined with multi-objective training that jointly optimizes mask generation, human–scene token-level prediction, and global prediction.

\begin{figure*}[!t]
    \centering
    \includegraphics[width=0.70\textwidth]{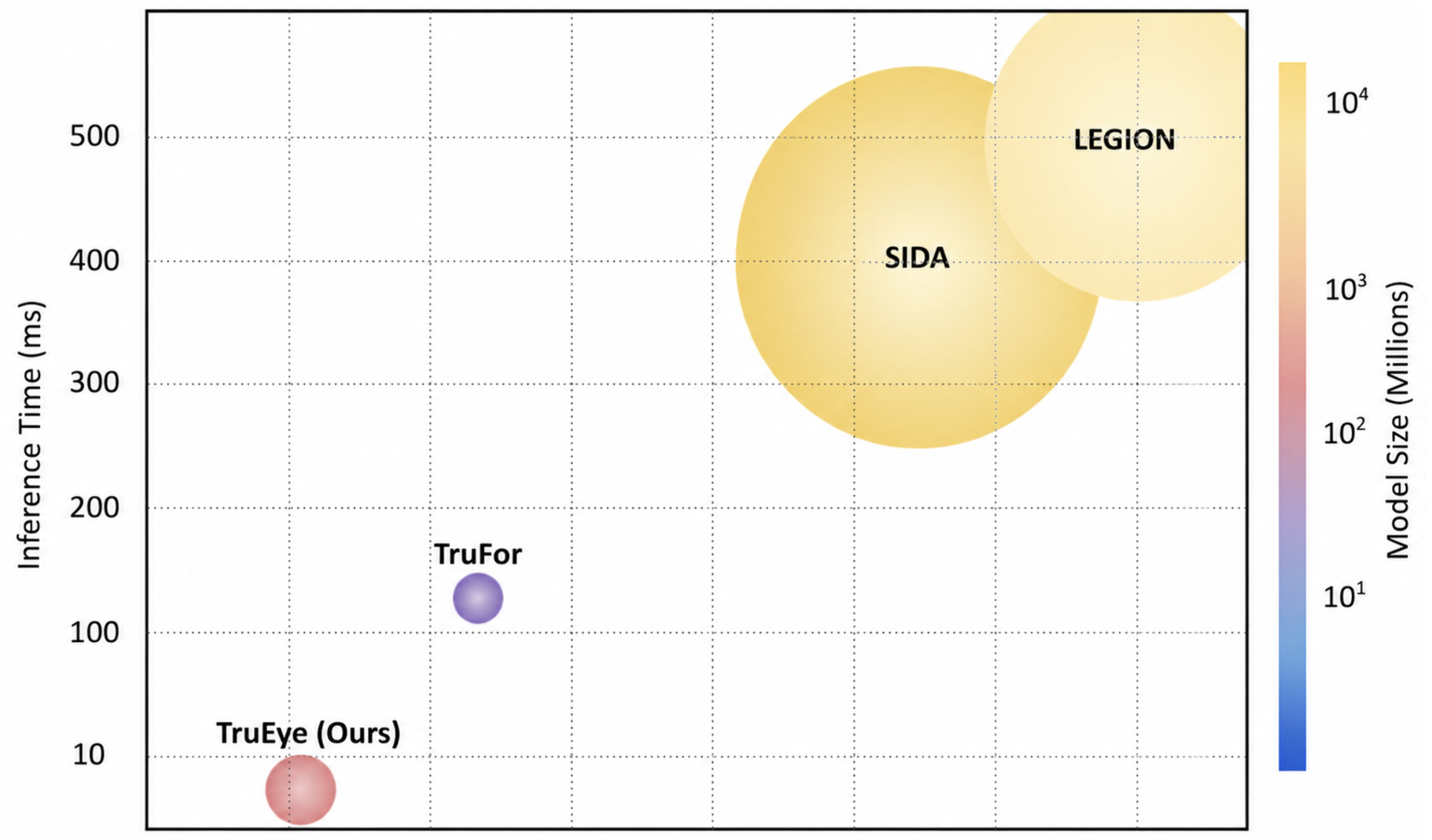}
        \caption{
       Inference Speed of Models 
        }

    \label{fig:time}
\end{figure*}

\subsubsection{Inference Speed}  

Figure \ref{fig:time} reports the inference time of all evaluated models, where the bubble size represents the corresponding model size. Each model is tested on 1,000 images using its official inference script, and the average per-image inference time is calculated.

As shown in the figure, our TruEye model achieves extremely low latency, i.e., only 4.57 ms per image. Its inference speed is comparable to the smallest models in the comparison, such as Adobe and XceptionNet. However, as discussed in the previous section, these smaller models exhibit significantly lower detection accuracy.

Compared with recent state-of-the-art models from 2025, including SIDA and LEGION, TruEye is over 100 times faster while also achieving higher detection accuracy. Specifically, SIDA requires an average of 460 ms per image, and LEGION requires 540 ms per image, yet both models still deliver lower detection accuracy than TruEye. In summary, TruEye achieves both superior accuracy and significantly lower latency with a much smaller parameter footprint, highlighting the efficiency of its architectural design.

\subsubsection{Robustness}

Table \ref{tab:robustness} reports performance under four aggressive perturbations: JPEG compression (QF=10), Gaussian blur ($\sigma=4$), downsampling to 0.3$\times$resolution, and additive Gaussian noise ($\sigma=0.1$). We randomly sampled 2,000 images (1,000 real and 1,000 synthetic) from the evaluation datasets, excluding FineSyn to ensure a fair out-of-distribution test. TruEye consistently outperforms the most recent baselines, SIDA and LEGION and TruFor  by a substantial margin.

\begin{table}[!ht]
\centering
\small
\setlength{\tabcolsep}{4pt}
\renewcommand{\arraystretch}{0.95}
\caption{Robustness Analysis under Different Perturbations}
\label{tab:robustness}
\resizebox{0.82\textwidth}{!}{%
\begin{tabular}{lcccccccc}
\toprule
\textbf{Type}
& \multicolumn{2}{c}{\textbf{TruEye (Ours)}}
& \multicolumn{2}{c}{\textbf{SIDA}}
& \multicolumn{2}{c}{\textbf{LEGION}}
& \multicolumn{2}{c}{\textbf{TruFor}} \\
\cmidrule(lr){2-3}
\cmidrule(lr){4-5}
\cmidrule(lr){6-7}
\cmidrule(lr){8-9}

& IoU(\%) & F1(\%)
& IoU(\%) & F1(\%)
& IoU(\%) & F1(\%)
& IoU(\%) & F1(\%) \\
\midrule

jpeg 10
& \textbf{89.97} & \textbf{91.05}
& 57.96 & 61.64
& 31.96 & 35.71
& 32.86 & 34.89 \\

resize 0.3
& \textbf{84.49} & \textbf{85.91}
& 66.50 & 68.66
& 31.28 & 35.15
& 33.67 & 34.22 \\

blur 4
& \textbf{80.28} & \textbf{80.90}
& 66.24 & 68.14
& 29.48 & 34.42
& 31.33 & 32.57 \\

noise 0.1
& \textbf{88.76} & \textbf{90.16}
& 60.45 & 62.92
& 30.96 & 35.57
& 32.51 & 32.87 \\

\bottomrule
\end{tabular}}

\end{table}

\vspace{-0.5em}

\subsection{Ablation Analysis}

In this ablation study, we analyze the contributions of the key components in TruEye including mask generation, dual-stream classification, feature magnification, and cross-attention mechanisms. Since the dual-stream classifier depends on the mask generator for stream separation and cannot operate independently, we focus on controlled variants that isolate the effects of dual-stream classification, feature magnification, and cross attention, respectively, to quantify their individual contributions to overall performance. Each variant is trained the same way as TruEye on the FineSyn dataset until the validation loss plateaus.

We first investigate whether the dual-stream architecture improves overall detection performance by comparing it with single-stream variants. Specifically, we consider two variants:
(i) V1: a plain Vision Transformer without the mask generator, which has been evaluated in Section 4.1 and demonstrates substantially lower detection accuracy. (ii) V2: a single-stream model equipped with the mask generator, where all image tokens are processed by a single stack of Transformer blocks followed by the feature magnification module, but without the cross-attention layer. As shown in the 1st row of Table~\ref{ablation_results}, the detection accuracy and F1 score drop significantly compared to the dual-stream architecture. This performance gap suggests that separating human and scene representations enables more specialized feature learning and reduces interference between heterogeneous visual cues. In contrast, a single-stream design forces the model to jointly encode human-specific artifacts and scene-level inconsistencies within a shared representation space, which may dilute discriminative signals.

Second, to assess the contribution of the feature magnification modules (FMM), we remove them from each Transformer block while keeping the remaining architecture unchanged.  As shown in 2nd row of Table~\ref{ablation_results}, eliminating the FMM leads to a clear drop in overall detection performance. This indicates that feature magnification enhances subtle discriminative cues, allowing the model to better amplify and capture fine-grained synthetic artifacts that may otherwise be suppressed during standard feature propagation.

Third, we evaluate the effectiveness of the cross-attention mechanism by removing the cross-attention computation and directly fusing the global [CLS] tokens from the two streams. As shown in the 3rd row of Table \ref{ablation_results}, when cross-attention is disabled, the detection accuracy of RHRS (Real Human Real Scene) drops most significantly. This result highlights the importance of the cross-attention mechanism in enhancing the model’s holistic understanding of the image.

\begin{table}[!t]
\centering
\caption{Ablation Analysis}
\label{ablation_results}
\resizebox{\columnwidth}{!}{%
\begin{tabular}{lcccccccccccccc}
\toprule
\textbf{Model Variant}
& \multicolumn{2}{c}{\textbf{SHSS}} 
& \multicolumn{2}{c}{\textbf{SHRS}} 
& \multicolumn{2}{c}{\textbf{SS}} 
& \multicolumn{2}{c}{\textbf{RHSS}} 
& \multicolumn{2}{c}{\textbf{RHRS}} 
& \multicolumn{2}{c}{\textbf{Real}} 
& \multicolumn{2}{c}{\bf Overall} \\
\cmidrule(lr){2-3} \cmidrule(lr){4-5} \cmidrule(lr){6-7}
\cmidrule(lr){8-9} \cmidrule(lr){10-11} \cmidrule(lr){12-13}
\cmidrule(lr){14-15}
& Acc(\%) & F1(\%) 
& Acc(\%) & F1(\%) & Acc(\%) & F1(\%) & Acc(\%) & F1(\%) & Acc(\%) & F1(\%) & Acc(\%) & F1(\%) & Acc(\%) & F1(\%) 
 \\
\midrule
Single Stream 
& -23.01 & -13.50 
& -23.90& -49.96 
& -8.90& -5.07 
& -39.85& -33.23 
& -11.02 & -6.06 
& -6.92 & -3.79 
& {\color{red}\underline{-18.93}} & {\color{red}\underline{-18.60}} \\

No FMM 
& -17.16 & -9.74 
& -6.23 & -10.51 
& -3.59 & -1.99 
& -10.88 & -8.51 
& -13.55 & -7.55 
& -9.08 & -5.03 
& -10.07 & -7.22 \\

No Cross Attention 
& -13.31 & -7.40 
& -6.75 & -11.09 
& -0.09 & -0.05 
& -5.36 & -4.03 
& {\color{red}\underline{-19.59}} & {\color{red}\underline{-11.29}} 
& -12.99 & -7.36 
& -9.68 & -6.86 \\

\bottomrule
\end{tabular}%
}
\end{table}

% ----  Linda: we will not report the following since we have sufficient other ablation analysis. 
%Finally, we explore model scaling by doubling the depth of the dual-stream classifier from 12 to 24 Transformer layers. Table~\ref{} shows that the deeper variant performs slightly worse than the 12-layer model. This degradation may stem from increased optimization difficulty in deeper networks and the reduced compatibility with pretrained 12-layer initialization.

\vspace{-1em}

\section{Conclusion}
\label{sec:conclusion}
In this work, we introduce TruEye, a novel framework for highly accurate fine-grained detection and localization of AI-generated or manipulated humans and scenes in images. TruEye has a unique mask-conditioned dual-stream transformer based architecture, which enables the model to capture subtle manipulations across both local and global contexts and to yield interpretable authenticity predictions that highlight where and what types of manipulations occur. The extensive experiments demonstrate that TruEye achieves consistently better generalization across diverse datasets  compared to state-of-the-art detectors. The newly curated FineSyn dataset further supports systematic fine-grained analysis of human–scene compositions for future research.  Overall, this work establishes a framework that strengthens robustness and trust in AI-generated image detection.

\bibliographystyle{splncs04}
\bibliography{main}
\end{document}